\pdfoutput=1
\documentclass[10pt,twocolumn]{article}

\usepackage[letterpaper,margin=0.75in]{geometry}
\usepackage[T1]{fontenc}
\usepackage[utf8]{inputenc}
\usepackage{lmodern}
\usepackage{microtype}

\usepackage{amsmath,amssymb,amsfonts,bm}
\usepackage{siunitx}

\usepackage{graphicx}
\usepackage{booktabs}
\usepackage{adjustbox}
\usepackage{algorithm}
\usepackage{algpseudocode}

\usepackage{xcolor}

\usepackage[hidelinks]{hyperref}
\usepackage[nameinlink,noabbrev]{cleveref}

\usepackage{authblk}

\usepackage[numbers,sort&compress]{natbib}

\title{Wavelet-Enhanced PaDiM for Industrial Anomaly Detection}

\author[1]{Cory Gardner}
\author[2]{Byungseok Min}
\author[1]{Tae-Hyuk Ahn}
\affil[1]{Department of Computer Science, Saint Louis University, St. Louis, MO 63103, USA\\
\texttt{\{cory.gardner, taehyuk.ahn\}@slu.edu}}
\affil[2]{Department of AI \& Data Science, Sejong University, Seoul 05006, South Korea\\
\texttt{bmin@sejong.ac.kr}}
\date{} % no date

\begin{document}
\maketitle

%====================
\begin{abstract}
Anomaly detection and localization in industrial images are essential for automated quality inspection. PaDiM, a prominent method, models the distribution of normal image features extracted by pre-trained Convolutional Neural Networks (CNNs) but typically relies on random channel selection for dimensionality reduction, potentially discarding structured information. We propose Wavelet-Enhanced PaDiM (WE-PaDiM), which integrates Discrete Wavelet Transform (DWT) analysis with multi-layer CNN features in a structured manner. WE-PaDiM applies 2D DWT individually to feature maps extracted from multiple layers of a backbone CNN. Specific frequency subband coefficients (e.g., approximation LL, details LH, HL) are selected from each layer's DWT output, spatially aligned if necessary, and then concatenated channel-wise before being modeled using PaDiM's patch-based multivariate Gaussian approach. This DWT-before-concatenation strategy provides a principled method for feature selection based on frequency content potentially relevant to anomalies, leveraging multi-scale information inherent in wavelet decomposition as an alternative to random selection. We evaluate WE-PaDiM on the challenging MVTec AD industrial anomaly dataset across multiple backbones (ResNet-18 and EfficientNet B0-B6). The proposed method achieves high performance in both anomaly detection and localization, yielding average results of approximately 99.32\% Image-AUC and 92.10\% Pixel-AUC across the 15 MVTec AD categories when using per-class optimized configurations. Our analysis suggests that specific wavelet configurations influence performance trade-offs; for instance, simpler wavelets (like Haar) combined with detail subbands (like HL or combinations including LH/HL/HH) often excel at precise localization, while configurations focusing on approximation bands (like LL) can enhance image-level detection. WE-PaDiM offers a competitive and interpretable alternative to random feature selection in PaDiM, achieving strong results suitable for industrial inspection tasks with comparable computational efficiency.
\end{abstract}


% ==============================
\section{Introduction}
\label{sec:introduction}
Anomaly detection in images, the process of identifying instances that deviate from established normal patterns, is a cornerstone technology for automated inspection in manufacturing and various industrial applications. Within this domain, the "one-class" learning setup is prevalent due to the frequent scarcity of anomalous data; models are trained exclusively on normal (defect-free) samples and tasked with detecting any unseen defects or deviations during inference.

Significant advancements have been made in unsupervised anomaly detection, particularly driven by benchmarks like the MVTec AD dataset \cite{bergmann2019mvtec}, which comprises 15 diverse industrial object and texture categories featuring realistic defect types. While early methods often relied on reconstruction techniques (e.g., autoencoders), recent embedding-based approaches leveraging features from pre-trained Convolutional Neural Networks (CNNs) have demonstrated superior performance. Among these, PaDiM (Patch Distribution Modeling) \cite{padim2020} emerged as an effective and computationally efficient technique. PaDiM models the statistical distribution of normal patch features, extracted from multiple layers of a pre-trained CNN, using multivariate Gaussian distributions to identify anomalies based on deviations from these learned norms.

A key aspect of the original PaDiM framework involves managing the high dimensionality that arises from concatenating feature vectors from multiple CNN layers. PaDiM addresses this by employing random feature selection, choosing a fixed-size subset of channels (e.g., 100 out of potentially 550+) before modeling the Gaussian distributions. While empirically effective and computationally beneficial compared to using all features or applying PCA, this random selection lacks structural basis and interpretability, potentially discarding channels or correlations valuable for identifying certain anomalies.

In this work, we propose Wavelet-Enhanced PaDiM (\textbf{WE-PaDiM}), modifying the feature processing stage to replace random selection with a structured, frequency-domain approach using the Discrete Wavelet Transform (DWT) \cite{mallat1989wavelet}. We hypothesize that anomalies in industrial images often manifest with specific frequency characteristics (e.g., fine scratches as high-frequency details, discolorations as low-frequency changes) that can be effectively captured by analyzing wavelet subbands of CNN features. Our method operates as follows: first, feature maps ($F_l$) are extracted from multiple pre-selected layers ($l \in L$) of a backbone CNN. Second, a 2D DWT is applied individually to each feature map $F_l$, decomposing it into different frequency subbands (e.g., LL, LH, HL, HH). Third, for each layer $l$, coefficients from a specific subset $S$ of these subbands (e.g., $F_{l,LL}, F_{l,LH}, F_{l,HL}$) are selected and then concatenated channel-wise, forming a per-layer wavelet feature map $\mathbf{f}_l$. Fourth, these per-layer wavelet feature maps $\mathbf{f}_l$ are spatially aligned (if necessary due to DWT downsampling) to a common resolution, resulting in $\mathbf{f}'_l$. Finally, these aligned per-layer maps $\mathbf{f}'_l$ are concatenated channel-wise across all selected layers $L$ to form the final feature vector $\mathbf{f}^{(w)}$ used for PaDiM's multivariate Gaussian modeling. This DWT-before-concatenation approach offers potential advantages: (1) it provides a principled, interpretable mechanism for feature selection and dimensionality management based on frequency content; (2) it allows leveraging multi-scale information captured by different subbands; and (3) it may enhance the separation between normal patterns and anomalies residing in specific frequency spectra.

We conducted a comprehensive evaluation of the proposed WE-PaDiM on the MVTec AD dataset, utilizing multiple pre-trained CNN backbones (including ResNet-18 \cite{he2016resnet} and variants of EfficientNet \cite{tan2019efficientnet}) to assess the method's generality. Our experiments demonstrate that WE-PaDiM achieves strong performance in both image-level anomaly detection and pixel-level anomaly localization, competitive with state-of-the-art methods. We perform ablation studies analyzing the impact of different wavelet families (e.g., Haar, Daubechies, Symlet) and various subband selection strategies (e.g., using only detail bands vs. including approximation bands). These analyses provide insights into which frequency components are most indicative of anomalies across different industrial contexts, revealing trade-offs between localization accuracy and detection sensitivity based on the chosen wavelet configuration. The WE-PaDiM approach maintains the inference efficiency characteristic of PaDiM, requiring only standard DWT operations and Gaussian modeling without iterative training or complex memory structures.

In summary, this paper makes the following contributions:
\begin{itemize}
\item We introduce Wavelet-Enhanced PaDiM (WE-PaDiM), a novel modification to the PaDiM framework where DWT is applied independently to multi-layer CNN features, followed by subband coefficient selection, alignment, and channel-wise concatenation prior to multivariate Gaussian modeling, serving as a structured alternative to random feature selection.
\item We rigorously designed and evaluated various WE-PaDiM configurations, exploring different wavelet types, subband selections, and CNN backbones, providing a detailed analysis of their influence on anomaly detection and localization performance on the MVTec AD benchmark.
\item We demonstrated that WE-PaDiM achieves high accuracy, performing competitively with existing state-of-the-art methods while retaining efficiency and offering potential gains in interpretability, achieving an average Image AUC of 99.32\% and an average Pixel AUC of 92.10\% on MVTec AD when using per-class optimized settings.
\item We provide insights derived from the wavelet-based analysis, particularly regarding the significance of specific frequency subbands for anomaly characterization and the trade-offs inherent in wavelet configuration choices, which may inform future research in deep anomaly detection.
\end{itemize}

The remainder of this paper is structured as follows: Section II reviews related work. Section III details the WE-PaDiM methodology. Section IV describes the experimental setup and presents the results. Section V discusses the findings and limitations. Section VI concludes the paper. Appendices provide detailed results.

%====================
\section{Related Work}
\label{sec:related}
\subsection{Unsupervised Anomaly Detection}
Early research on visual anomaly detection often employed reconstruction-based methods. For example, Bergmann et al. introduced the MVTec AD dataset and evaluated classical methods such as autoencoders and variational autoencoders (VAE) for unsupervised anomaly detection \cite{bergmann2019mvtec}. Reconstruction approaches train on normal images and attempt to detect anomalies by high reconstruction error. 
Over time, various improvements were explored: e.g. using structural similarity (SSIM) loss to better highlight defects in reconstructions \cite{bergmann2018ssimAE}, memory-augmented autoencoders to avoid learning abnormal features \cite{gong2019memae}, and 
generative adversarial network(GAN) models to constrain the latent space of normal images \cite{schlegl2017anogan, akcay2018ganomaly}. Despite these advances, reconstruction methods often struggled with fine defects or in producing accurate pixel-level localization.
Recent approaches have shifted to embedding-based anomaly detection, which leverages rich features from pre-trained deep networks (often CNNs trained on ImageNet) and does not require reconstructing the input.   %corrected(Min)
For instance, SPADE (Sub-image Anomaly Detection) by Cohen et al. used features from a Wide ResNet-50 and a spatial pyramid to detect anomalies without reconstructive models \cite{cohen2020spade}. PatchCore \cite{roth2022patchcore} further improved performance by using a memory bank of normal patch features and an efficient subsampling strategy to perform nearest-neighbor based anomaly scoring, achieving nearly perfect detection on MVTec AD. 
%\cite{bergmann2020studentteacher, li2021cutpaste, yu2023fastflow, zavrtanik2021draem, ruff2021differnet}. 
Other notable contributions include the student–teacher distillation approach of Bergmann et al.\cite{bergmann2020studentteacher}, the CutPaste self-supervised augmentation scheme of Li et al.\cite{li2021cutpaste}, the real-time FastFlow density estimator of Yu et al.\cite{yu2023fastflow}, the hybrid inpainting-based DRAEM framework of Zavrtanik et al.\cite{zavrtanik2021draem}, and DifferNet’s normalizing-flow modeling of Ruff et al.~\cite{ruff2021differnet},.
All of these show the strength of using rich pre-trained features and feature selection in anomaly detection.

\subsection{PaDiM Framework}
Defard et al. proposed PaDiM as a fast yet powerful anomaly detection and localization method \cite{padim2020}. PaDiM uses a pre-trained CNN (e.g., ResNet18 \cite{he2016resnet}) and extracts multiple layers' feature maps for an input image. At each spatial location (patch), the feature vectors (concatenated from selected layers) are modeled as a multivariate Gaussian distribution (with mean and covariance) estimated from normal training images. At test time, an anomaly score per location is computed via the Mahalanobis distance of the patch feature to the normal distribution; an image-level anomaly score is obtained by taking the maximum patch score. PaDiM demonstrated that modeling the correlations between feature dimensions (i.e., using the full covariance matrix rather than independent assumptions) improves localization accuracy. To keep the covariance manageable, PaDiM randomly selects a subset of feature dimensions (for example, 100) out of the full set (which may be several hundred channels across layers). An ablation in the PaDiM paper showed that this random selection causes negligible performance drop compared to using all features, while dramatically reducing memory and computation. Surprisingly, random selection performed better than PCA-based selection, possibly because PCA prioritizes high-variance principal components which might correspond to global structure, while anomalies may be in fine-detail components. Since its release, several variants of PaDiM have appeared. Light PaDiM \cite{ren2022lightpadim} compresses the original framework by pruning redundant channels and simplifying covariance estimation, achieving comparable accuracy on MVTec AD while reducing memory and inference time for deployment on resource-limited devices. More recently, PaDiM-ACE \cite{ibarra2025padimace} substitutes the unbounded Mahalanobis distance with an Adaptive Cosine Estimator similarity, yielding bounded, more robust anomaly scores and demonstrating improved performance on challenging synthetic-aperture-radar imagery.

\begin{figure*}[h!]
    \centering
    \includegraphics[width=0.95\textwidth]{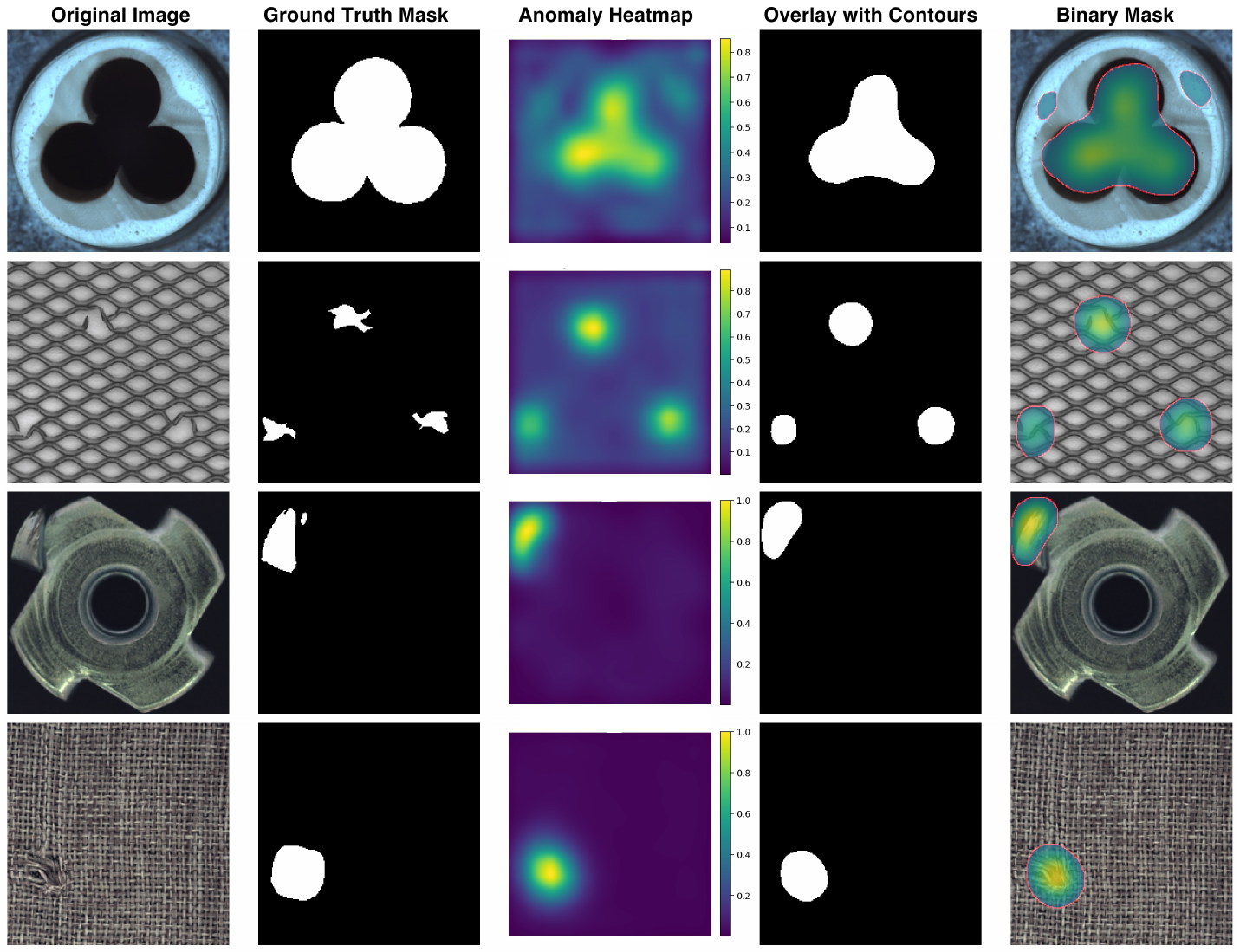} 
    \caption{Visualization of anomaly detection results using our WE-PaDiM method on the MVTec AD dataset. From left to right: (1) Original input images containing anomalies, (2) Ground truth anomaly masks highlighting the actual defects, (3) Generated anomaly heatmaps where yellow/green regions indicate high anomaly scores and purple regions represent normal areas, (4) Overlay of detected anomaly contours on the ground truth segmentation, and (5) Final binary anomaly masks with colorized heatmap overlay on the original image. The examples show various industrial defect types: anomaly in the cross-section of a cable (row 1), structural anomalies in a mesh/grid (row 2), broken corner in a metal nut (row 3), and a small hole in fabric (row 4). Our method effectively localizes anomalies of different sizes and shapes across various textures and object categories, demonstrating strong pixel-level segmentation performance.}
    \label{fig:architecture_workflow}
\end{figure*}

\subsection{Wavelets in Computer Vision}
The Discrete Wavelet Transform (DWT) provides a multi-scale decomposition of images into subband frequencies (typically labeled LL for low-frequency approximation, LH for horizontal detail, HL for vertical detail, and HH for diagonal detail). Wavelets have been used in classical computer vision for texture analysis and noise filtering \cite{mallat1989wavelet}, and more recently in deep learning to incorporate multi-resolution analysis. For example, wavelet scattering transforms \cite{mallat2012wavelet} have been proposed as a fixed feature extractor that is stable to deformations and preserves high-frequency information for tasks like classification. In anomaly detection, wavelets have been applied mostly in time-series analysis \cite{kanarachos2015anomaly} or as a preprocessing step for denoising \cite{donoho1995adapting}. In the context of deep anomaly detection, our work is novel in that it integrates wavelet decomposition with deep feature embeddings during the anomaly feature modeling stage. We aim to capture anomalies at different scales by decomposing CNN feature maps. Generally, the LL subband retains the coarse image layout (potentially useful for detecting large structural anomalies or overall intensity shifts) while LH/HL/HH subbands are associated with finer textures and edges (potentially capturing defects like small scratches or cracks). We experiment with different wavelet types including Haar (a simple 1-level Daubechies wavelet) and more complex wavelets like Daubechies-2 (db2) and Symlet-4 (sym4), which have longer filters and can capture more contextual information in the subbands.

To our knowledge, no prior work has explicitly replaced learned or random feature selection in deep anomaly detection with a wavelet-based selection. By evaluating wavelet-enhanced features within the PaDiM framework, we contribute a perspective on how frequency-domain analysis can add to modern anomaly detection methods.

%====================
\section{Methodology: Wavelet-Enhanced PaDiM}
\label{sec:method}

Our proposed method, Wavelet-Enhanced PaDiM (WE-PaDiM), builds upon the PaDiM framework by incorporating Discrete Wavelet Transform (DWT) analysis \emph{prior} to modeling the distribution of normal features. Instead of concatenating raw CNN features and then applying random selection, we first decompose features from multiple layers using DWT, select informative subband coefficients, align them spatially, and then concatenate these coefficients channel-wise before applying PaDiM's core modeling technique.

%====================
\subsection{Review of PaDiM}
\label{subsec:padim_review}

We first briefly summarize the original PaDiM method \cite{padim2020} to provide context. PaDiM leverages feature embeddings from a CNN backbone pre-trained on a large dataset like ImageNet. Let $f_l(\cdot)$ denote the function extracting the feature map $F_l \in \mathbb{R}^{C_l \times H_l \times W_l}$ from a chosen layer $l$, where $C_l$ is the channel count and $(H_l, W_l)$ are the spatial dimensions. PaDiM typically selects a set of layers $L$ (e.g., for ResNet-18 \cite{he2016resnet}, the outputs of \texttt{layer1}, \texttt{layer2}, \texttt{layer3}; for EfficientNets \cite{tan2019efficientnet}, intermediate blocks like \texttt{features.3} through \texttt{features.6}.
) to capture multi-level information.

In the original PaDiM, these feature maps $F_l$ are spatially aligned (usually by resizing later-layer features to match the earliest layer's size) and concatenated channel-wise at each location $(x,y)$ to form a combined feature vector $f(x,y) \in \mathbb{R}^D$, where $D = \sum_{l \in L} C_l$. To manage the potentially large dimension $D$, PaDiM performs dimensionality reduction by randomly selecting a subset of $D_R$ dimensions (e.g., $D_R=100$ or $D_R=550$) from $f(x,y)$, yielding $f^{(R)}(x,y) \in \mathbb{R}^{D_R}$.

During training on $N$ normal images $\{I_n\}_{n=1}^N$, PaDiM estimates the parameters of a Multivariate Gaussian (MVG) distribution for the selected feature vectors at each spatial location $(x,y)$. Specifically, it calculates the sample mean $\hat{\bm{\mu}}^{(R)}_{x,y} \in \mathbb{R}^{D_R}$ and the regularized sample covariance matrix $\hat{\bm{\Sigma}}^{(R)}_{x,y} \in \mathbb{R}^{D_R \times D_R}$ using the features $\{f^{(R)}_n(x,y)\}_{n=1}^N$.

At inference time, for a test image $I_{test}$, the corresponding feature vector $f^{(R)}_{test}(x,y)$ is extracted. Its Mahalanobis distance to the learned normal distribution at that location serves as the anomaly score:
\begin{multline}
s^{(R)}(x,y) = \\
\sqrt{(f^{(R)}_{test}(x,y) - \hat{\bm{\mu}}^{(R)}_{x,y})^T (\hat{\bm{\Sigma}}^{(R)}_{x,y})^{-1} (f^{(R)}_{test}(x,y) - \hat{\bm{\mu}}^{(R)}_{x,y})}.
\end{multline}
The collection of these scores forms an anomaly map $S^{(R)}$, and the maximum score across the map is often used as the image-level anomaly score (Figure 1).

\subsection{WE-PaDiM: Feature Extraction and Per-Layer DWT}
\label{subsec:wepadim_dwt}

WE-PaDiM modifies the feature processing stage. Like PaDiM, it begins by extracting feature maps $F_l \in \mathbb{R}^{C_l \times H_l \times W_l}$ for each selected layer $l \in L$ from the pre-trained CNN backbone using input image $I$.

The core difference lies in the next step: instead of immediate concatenation or random selection, we apply a 2D Discrete Wavelet Transform (DWT) individually to each feature map $F_l$. We utilize the pytorch-wavelets library for efficient DWT computation. For a chosen wavelet basis $\psi$ (e.g., Haar, db4, sym4) and decomposition level $J$ (1 or 2 in our experiments), the DWT decomposes each channel $c$ of $F_l$ into subbands. For $J=1$, this yields four sets of coefficients:
\begin{itemize}
    \item $F^{LL}_l$: Approximation coefficients (Low-Low frequencies).
    \item $F^{LH}_l$: Horizontal detail coefficients (Low-High frequencies).
    \item $F^{HL}_l$: Vertical detail coefficients (High-Low frequencies).
    \item $F^{HH}_l$: Diagonal detail coefficients (High-High frequencies).
\end{itemize}
For an orthogonal wavelet like Haar or Daubechies and $J=1$, each subband $F^s_l$ (where $s \in \{LL, LH, HL, HH\}$) has dimensions $C_l \times H'_l \times W'_l$, where $H'_l \approx H_l / 2^J$ and $W'_l \approx W_l / 2^J$. If $J > 1$, the decomposition is applied recursively to the approximation coefficients $F^{LL}_l$ of the previous level ($J-1$).

%====================
\subsection{WE-PaDiM: Subband Selection, Alignment, and Concatenation}
\label{subsec:wepadim_select_align_concat}

After obtaining the DWT coefficients for each layer, we perform three steps to construct the final feature map $\bm{f}^{(w)}$ used for modeling:

\begin{enumerate}
    \item \textbf{Subband Selection:} We define a subset $S \subseteq \{LL, LH, HL, HH\}$ of subbands to retain. This allows us to focus on specific frequency information and potentially reduce dimensionality if $|S| < 4$. For example, $S = \{LL, LH, HL\}$ excludes the diagonal details often associated with noise.

    \item \textbf{Per-Layer Coefficient Concatenation:} For \emph{each layer} $l \in L$, we concatenate the $C_l$ channels from \emph{all selected subbands} $s \in S$ channel-wise. This results in an intermediate feature map $\bm{f}_l$ for each layer:
    \begin{equation}
        \bm{f}_l = \text{Concat}_{s \in S}(\bm{F}^s_l) \in \mathbb{R}^{(|S| \cdot C_l) \times H'_l \times W'_l}.
        \label{eq:per_layer_concat}
    \end{equation}

    \item \textbf{Spatial Alignment and Final Concatenation:} The spatial dimensions $(H'_l, W'_l)$ of the resulting $\bm{f}_l$ maps might differ across layers $l \in L$. This occurs if the original feature maps $(H_l, W_l)$ had different sizes, or potentially if different DWT levels $J$ were used per layer (though typically $J=1$ is used for all). To enable concatenation for PaDiM's patch-based modeling, these maps must be aligned to a common spatial resolution $(H', W')$. The target size $(H', W')$ is typically chosen as the spatial dimension of the DWT coefficients corresponding to the earliest layer $l \in L$ (i.e., the layer with the largest $H_l, W_l$). Let $\bm{f}'_l \in \mathbb{R}^{(|S| \cdot C_l) \times H' \times W'}$ denote the spatially aligned feature map for layer $l$. The final wavelet-enhanced feature map $\bm{f}^{(w)}$ is obtained by concatenating these aligned maps channel-wise across all layers:
    \begin{equation}
    \label{eq:final_concat}
        \bm{f}^{(w)} = \text{Concat}_{l \in L}(\bm{f}'_l) \in \mathbb{R}^{D_W \times H' \times W'}.
    \end{equation}
\end{enumerate}
The total channel dimension $D_W$ of this final map is the sum of the channel dimensions from the selected subbands across all layers:
\begin{equation}
    D_W = \sum_{l \in L} |S| \cdot C_l.
\end{equation}
The final feature dimension $D_W$ depends directly on the number of layers $|L|$, the number of selected subbands $|S|$, and the original channel counts $C_l$ of the selected layers. Dimensionality reduction compared to concatenating the raw features (dimension $D = \sum C_l$) occurs primarily if $|S| < 4$. Compared to original PaDiM's random selection to dimension $D_R$, $D_W$ may be larger or smaller, but represents a structured, frequency-informed feature set.

%====================
\subsection{WE-PaDiM: Anomaly Modeling and Scoring}
\label{subsec:wepadim_model_score}

Given the final wavelet-enhanced feature map $\bm{f}^{(w)} \in \mathbb{R}^{D_W \times H' \times W'}$ obtained via Eq. \ref{eq:final_concat}, we proceed with PaDiM's core modeling strategy.

During training on $N$ normal images $\{I_n\}_{n=1}^N$, we collect the corresponding feature maps $\{\bm{f}^{(w)}_n\}_{n=1}^N$. For each spatial location $(i,j)$ in the $H' \times W'$ grid ($1 \le i \le H'$, $1 \le j \le W'$), we estimate the parameters of a Multivariate Gaussian distribution $\mathcal{N}(\bm{\mu}^{(w)}_{i,j}, \bm{\Sigma}^{(w)}_{i,j})$ that models the distribution of the $D_W$-dimensional vectors $\{\bm{f}^{(w)}_n(i,j)\}_{n=1}^N$. The sample mean $\hat{\bm{\mu}}^{(w)}_{i,j}$ and the sample covariance matrix $\hat{\bm{C}}^{(w)}_{i,j}$ are calculated:
\begin{align}
    \hat{\bm{\mu}}^{(w)}_{i,j} &= \frac{1}{N} \sum_{n=1}^N \bm{f}^{(w)}_n(i,j) \\
    \hat{\bm{C}}^{(w)}_{i,j} &= \frac{1}{N-1} \sum_{n=1}^N (\bm{f}^{(w)}_n(i,j) - \hat{\bm{\mu}}^{(w)}_{i,j})(\bm{f}^{(w)}_n(i,j) - \hat{\bm{\mu}}^{(w)}_{i,j})^T
\end{align}
We compute and store these statistics \textbf{per spatial location $(i,j)$}. To ensure the covariance matrix is invertible and numerically stable, we add a small regularization term $\epsilon$. 
) to its diagonal:
\begin{equation}
    \hat{\bm{\Sigma}}^{(w)}_{i,j} = \hat{\bm{C}}^{(w)}_{i,j} + \epsilon \bm{I},
\end{equation}
where $\bm{I}$ is the $D_W \times D_W$ identity matrix. The parameters $\{\hat{\bm{\mu}}^{(w)}_{i,j}, \hat{\bm{\Sigma}}^{(w)}_{i,j}\}$ for all $(i,j)$ constitute the learned model of normality.

At inference time, given a test image $I_{test}$, we compute its corresponding wavelet-enhanced feature map $\bm{f}^{(w)}_{test}$. Then, for each location $(i,j)$, we calculate the squared Mahalanobis distance to the learned normal distribution at that location.
\begin{equation}
    s_w(i,j) = (\bm{f}^{(w)}_{test}(i,j) - \hat{\bm{\mu}}^{(w)}_{i,j})^T (\hat{\bm{\Sigma}}^{(w)}_{i,j})^{-1} (\bm{f}^{(w)}_{test}(i,j) - \hat{\bm{\mu}}^{(w)}_{i,j}).
    \label{eq:mahalanobis}
\end{equation}
This calculation uses the pre-computed inverse covariance. The resulting values $s_w(i,j)$ form the raw anomaly map $S_w \in \mathbb{R}^{H' \times W'}$.

This anomaly map $S_w$ is typically post-processed:
\begin{enumerate}
    \item \textbf{Upsampling:} Resized to match the original input image resolution (e.g., $224 \times 224$) using bilinear interpolation.
    \item \textbf{Smoothing:} Optionally smoothed using a Gaussian filter to reduce noise and enhance spatially coherent anomaly regions.
\end{enumerate}
The final image-level anomaly score $s_{w,img}$ is derived from the post-processed anomaly map, typically by taking the maximum value across all pixels: $s_{w,img} = \max_{i,j} s_w(i,j)$.

%====================
\subsection{Algorithm Summary}
\label{subsec:wepadim_algo}
Algorithm \ref{alg:wepadim} summarizes the training and inference phases of the proposed WE-PaDiM method based on the DWT-before-concatenation approach.

\begin{algorithm}[h!]
\footnotesize
\caption{Wavelet‐Enhanced PaDiM (WE‐PaDiM)}
\label{alg:wepadim}
\begin{algorithmic}[1]

\Require Normal training images $\{I_n\}_{n=1}^{N}$; test image $I_{\text{test}}$; CNN backbone $f$; selected layers $L$; wavelet type $\psi$; level $J$; subband set $S$; target size $(H',W')$; regularization $\epsilon$.
\Ensure  Anomaly map $S_w$ and image score $s_{w,\text{img}}$ for $I_{\text{test}}$

\State \(\triangleright\)\, \textbf{Training phase}

\State Initialise feature sets $\mathcal F_{i,j}\gets\varnothing$ for all $(i,j)$ on the $H'\!\times\!W'$ grid
\For{$n=1$ \textbf{to} $N$}                                   \Comment{normal training images}
    \State Extract $\{F_l=f_l(I_n)\in\mathbb R^{C_l\times H_l\times W_l}\}_{l\in L}$
    \State $A_n\gets[\;]$ \Comment{aligned layer features list}
    \For{$l\in L$}                                             \Comment{per layer}
        \State $(Y_{L,l},Y_{H,l})\gets\mathrm{DWT}_{\psi,J}(F_l)$
        \State $\{F_l^{s}\}_{s\in S}\gets$ selected subbands from $(Y_{L,l},Y_{H,l})$
        \State $\mathbf f_l\gets\text{Concat}_{s\in S}(F_l^{s})$
        \State $\mathbf f'_l\gets\text{Interpolate}(\mathbf f_l,\text{size}=(H',W'))$
        \State Append $\mathbf f'_l$ to $A_n$
    \EndFor
    \State $\mathbf f^{(w)}_n\gets\text{Concat}_{l\in L}(A_n)\in\mathbb R^{D_W\times H'\times W'}$
    \ForAll{locations $(i,j)$}
        \State Add $\mathbf f^{(w)}_n(i,j)$ to $\mathcal F_{i,j}$
    \EndFor
\EndFor
\ForAll{locations $(i,j)$}                                     \Comment{compute statistics}
    \State $\hat{\boldsymbol\mu}^{(w)}_{i,j}\gets\text{mean}(\mathcal F_{i,j})$
    \State $\hat{\mathbf C}^{(w)}_{i,j}\gets\text{cov}(\mathcal F_{i,j})$
    \State $\hat{\boldsymbol\Sigma}^{(w)}_{i,j}\gets\hat{\mathbf C}^{(w)}_{i,j}+\epsilon\mathbf I$
    \State Store $(\hat{\boldsymbol\Sigma}^{(w)}_{i,j})^{-1}$
\EndFor

\vspace{0.5ex}

\State \(\triangleright\)\, \textbf{Inference phase}

\State Extract $\{F_l=f_l(I_{\text{test}})\}_{l\in L}$, build $A_{\text{test}}$ exactly as above
\State $\mathbf f^{(w)}_{\text{test}}\gets\text{Concat}_{l\in L}(A_{\text{test}})$
\State Initialise $S_w\in\mathbb R^{H'\times W'}$
\ForAll{locations $(i,j)$}                                     \Comment{Mahalanobis distance}
    \State $S_w(i,j)\gets(\mathbf f^{(w)}_{\text{test}}(i,j)-\hat{\boldsymbol\mu}^{(w)}_{i,j})^{\!\top}
           (\hat{\boldsymbol\Sigma}^{(w)}_{i,j})^{-1}
           (\mathbf f^{(w)}_{\text{test}}(i,j)-\hat{\boldsymbol\mu}^{(w)}_{i,j})$
\EndFor
\State Upsample $S_w$ to the original image size and optionally smooth
\State $s_{w,\text{img}}\gets\max_{i,j} S_w(i,j)$
\Return $S_w,\; s_{w,\text{img}}$

\end{algorithmic}
\end{algorithm}

%====================
\subsection{Complexity and Implementation}
\label{subsec:wepadim_complexity}

The WE-PaDiM approach maintains efficiency comparable to the original PaDiM.
\begin{itemize}
    \item \textbf{DWT Cost:} Applying a 2D DWT per layer using efficient libraries like pytorch-wavelets adds minimal computational overhead compared to the initial CNN feature extraction pass. The complexity is approximately linear in the number of pixels in the feature maps being transformed.
    \item \textbf{Alignment Cost:} The spatial alignment using interpolation (e.g., bilinear) is also computationally inexpensive.
    \item \textbf{Statistics Storage:} The primary memory requirement stems from storing the mean vectors and covariance matrices. As we compute and store statistics \textbf{per spatial location} $(i,j)$, the total storage complexity is dominated by the covariance matrices, scaling as $O(P \cdot D_W^2)$, where $P=H' \cdot W'$ is the number of spatial locations in the final feature map $\bm{f}^{(w)}$ and $D_W$ is its channel dimensionality (Eq. \ref{eq:final_concat}). While $D_W$ can be large (potentially >1000 depending on the configuration), $P$ is often relatively small (e.g., $14 \times 14=196$). The memory-efficient batch processing implemented for covariance calculation helps manage peak usage during training.
    \item \textbf{Inference Cost:} The inference time is primarily determined by the CNN forward pass, the DWT/alignment/concatenation steps, and the Mahalanobis distance calculation. The Mahalanobis computation involves approximately $O(P \cdot D_W^2)$ operations per image (assuming pre-computed inverses). Overall, the process remains efficient, suitable for many industrial applications, especially when leveraging GPU acceleration.
\end{itemize}
Compared to the original PaDiM using a randomly selected subset of $D_R$ dimensions, WE-PaDiM's complexity scales with the structured dimension $D_W$. Choosing $|S| < 4$ can reduce $D_W$, potentially lowering computational and memory requirements compared to using all raw features ($D$), while providing a more interpretable feature set than random selection to $D_R$.

\begin{figure*}[h!]
    \centering
    \includegraphics[width=0.95\textwidth]{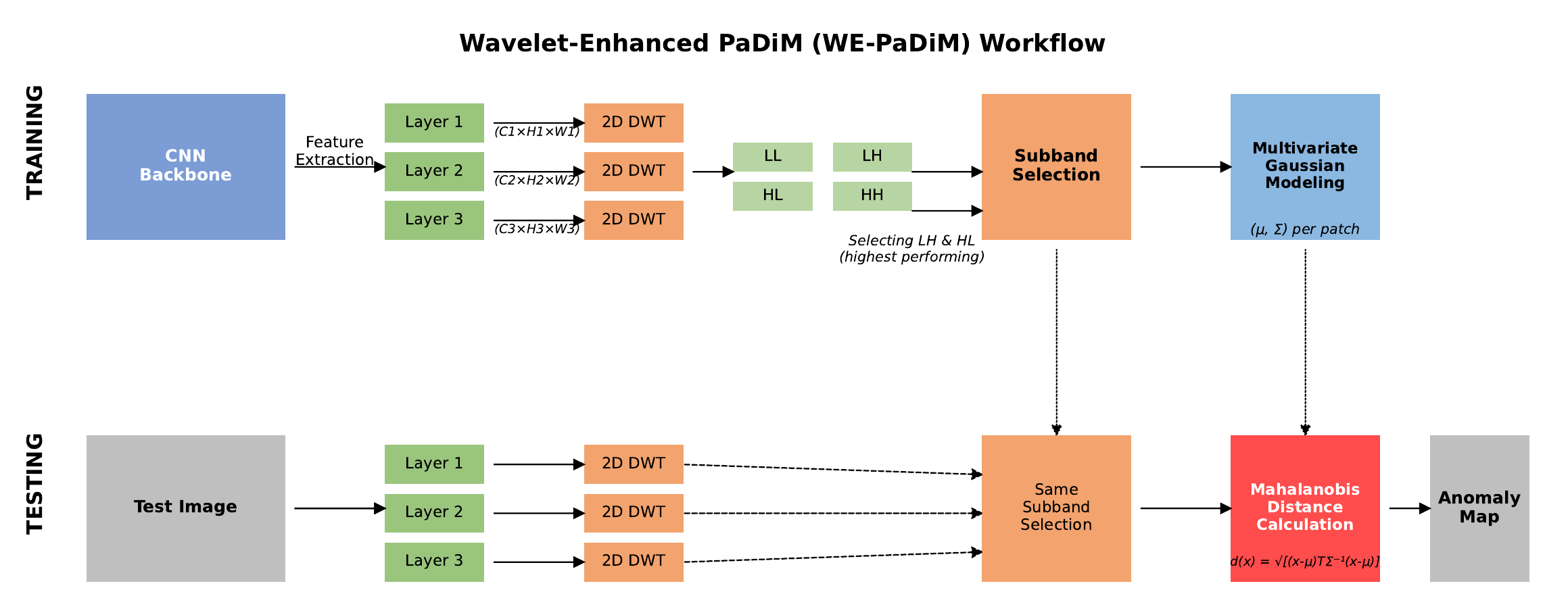} 
    \caption{Overview of the Wavelet-Enhanced PaDiM (WE-PaDiM) architecture. The diagram illustrates the WE-PaDiM workflow for both training (top path) and inference (bottom path). In the training phase, normal input images are processed by a CNN backbone to extract feature maps ($F_l$) from multiple layers (e.g., Layer 1, 2, 3 with dimensions $C_l \times H_l \times W_l$). Each layer's feature map undergoes an individual 2D Discrete Wavelet Transform (DWT), decomposing it into subbands (e.g., LL, LH, HL, HH). Following subband selection (e.g., LH \& HL), the chosen wavelet coefficients from each layer are spatially aligned and then concatenated channel-wise across layers to form the final feature representation ($\mathbf{f}^{(w)}$). These features are used to learn patch-wise multivariate Gaussian distributions ($\bm{\mu}^{(w)}, \bm{\Sigma}^{(w)}$) modeling normality. During inference, a test image undergoes the identical feature extraction and "DWT-before-concatenation" processing. The resulting features are then compared against the learned normal distributions using Mahalanobis distance to generate an anomaly map. This structured, frequency-domain feature processing is a key differentiator of WE-PaDiM.}
    \label{fig:architecture_workflow}
\end{figure*}
%====================

%====================
\section{Experiments}
\label{sec:experiments}

This section details the experimental setup used to evaluate the proposed Wavelet-Enhanced PaDiM (WE-PaDiM) method, including the dataset, evaluation metrics, implementation specifics, and the structure for presenting results.

%====================
\subsection{Dataset and Evaluation Metrics}
\label{subsec:dataset_metrics}

We evaluate our method on the widely used \textbf{MVTec Anomaly Detection (MVTec AD) dataset} \cite{bergmann2019mvtec}. This dataset is specifically designed for unsupervised anomaly detection in industrial settings and comprises 15 different categories, including 5 texture types and 10 object types. Each category contains a set of normal (defect-free) images designated for training and a separate test set containing both normal images and images with various types of anomalies. Crucially, the test set provides pixel-level ground truth masks for the anomalous regions.

Following the standard one-class learning protocol for MVTec AD, we train a separate model for each of the 15 categories using only the normal images from the respective training set. Evaluation is then performed on the test set of each category.

We report performance using three standard metrics, calculated per category and then averaged across all 15 categories:

\begin{itemize}
    \item \textbf{Image-level ROC AUC (Image AUC)}: This metric evaluates the model's ability to classify entire images as either normal or anomalous. It is calculated by computing the Area Under the Receiver Operating Characteristic Curve based on the image-level anomaly scores ($s_{w,img}$) derived for each test image against their ground truth labels (0 for normal, 1 for anomalous). An AUC of 1.0 represents perfect classification, while 0.5 indicates random chance performance.

    \item \textbf{Pixel-level ROC AUC (Pixel AUC)}: This metric assesses the model's localization accuracy by evaluating its ability to classify individual pixels as normal or anomalous. The pixel-wise anomaly scores from the final (upsampled and smoothed) anomaly map $S_w$ are compared against the ground truth binary masks provided by the dataset. The AUC is computed across all pixels from all test images within a category.

\end{itemize}

%====================
\subsection{Implementation Details}
\label{subsec:implementation}

Our WE-PaDiM implementation builds upon PyTorch \cite{paszke2019pytorch} and integrates the pytorch-wavelets library for DWT computations.

\textbf{Backbone CNNs}: We utilized pre-trained CNNs from torchvision.models as feature extractors. To evaluate robustness, we experimented with ResNet-18, and EfficientNet variants (B0, B1, B2, B3, B4, B5, B6). All models were initialized with weights pre-trained on ImageNet and used without any fine-tuning on MVTec AD data.

\textbf{Feature Extraction}: Following common practice in embedding-based anomaly detection, we extracted features from multiple layers to capture varying levels of abstraction.
\begin{itemize}
    \item For ResNet backbones (e.g., ResNet-18), features were extracted from the output of \texttt{layer1}, \texttt{layer2}, and \texttt{layer3}.
    \item For EfficientNet backbones (e.g., B0-B6), features were extracted from intermediate blocks within the main \texttt{features} sequence, specifically targeting the outputs corresponding to indices 3, 4, 5, and 6, representing increasing depths.
\end{itemize}

\textbf{WE-PaDiM Configuration}: The core of our method involves the DWT-before-concatenation process described in Section \ref{sec:method}.
\begin{itemize}
    \item \textbf{DWT:} We primarily used a single-level ($J=1$) 2D DWT implemented via pytorch-wavelets. We explored different wavelet families, including Haar, Daubechies (db2, db4), and Symlet (sym4) during our parameter search. Some ablation studies also considered $J=2$.
    \item \textbf{Subband Selection:} For each layer, coefficients corresponding to a selected subset $S$ of subbands were concatenated channel-wise. We systematically evaluated different combinations of $S$, including individual subbands and various multi-subband sets (e.g., $\{LL, LH, HL\}$, all 15 non-empty combinations) during different experimental stages.
    \item \textbf{Alignment \& Concatenation:} The DWT coefficient maps from different layers were spatially aligned to a common resolution (determined by the earliest layer's post-DWT size) using bilinear interpolation (\texttt{torch interpolate}) before being concatenated channel-wise across layers to form the final feature map $\bm{f}^{(w)}$.
\end{itemize}

\textbf{Parameter Ranges \& Training}: Key hyperparameters were explored using grid search:
\begin{itemize}
    \item Gaussian smoothing $\sigma$: Explored values such as 2.0, 4.0, and 6.0.
    \item Covariance regularization $\epsilon$: Explored values such as 0.1, 0.01, and 0.001.
\end{itemize}
For each specific configuration (backbone, class, wavelet type, level, subbands, sigma, cov\_reg), the model statistics ($\hat{\bm{\mu}}^{(w)}_{i,j}$, $\hat{\bm{\Sigma}}^{(w)}_{i,j}$) were computed using features from all available normal training images for that class. We compute and store statistics \textbf{per spatial location} and apply regularization $\epsilon \bm{I}$ to the sample covariance matrix $\hat{\bm{C}}^{(w)}_{i,j}$.

\textbf{Inference \& Scoring}: During testing, features $\bm{f}^{(w)}_{test}$ are extracted and processed identically. The Mahalanobis distance $s_w(i,j)$ is computed at each location using the learned statistics. The resulting anomaly map $S_w$ is upsampled to the original input image size (typically $224 \times 224$ after center cropping from a $256 \times 256$ resize) and smoothed with a Gaussian filter using the specified $\sigma$. The final image-level anomaly score $s_{w,img}$ is the maximum value of the smoothed map.

\textbf{Testing Hardware}: Experiments were conducted using an NVIDIA A100 GPU (40GB). 

%====================
\subsection{Overall Performance}
\label{subsec:results_overall}

WE-PaDiM demonstrates strong overall performance on the MVTec AD dataset. When optimized for Image AUC, the average performance across all 15 classes, using the best consolidated configuration for each class, reached \textbf{99.32\%}. Similarly, when optimized for Pixel AUC, the average performance was \textbf{92.10\%}.

Table \ref{tab:best_avg_image_auc_backbone} presents the best average Image AUC achieved by each backbone model across all classes, along with the corresponding configuration. EfficientNet-B6 yielded the highest average Image AUC of 0.9884. In contrast, Table \ref{tab:best_avg_pixel_auc_backbone} shows that ResNet-18 achieved the best average Pixel AUC of 0.9077. These results highlight that different backbones and wavelet configurations may be optimal depending on whether image-level detection or pixel-level localization is prioritized.

\begin{table*}[h!]
\centering
\caption{Best average image AUC configuration by backbone model.}
\label{tab:best_avg_image_auc_backbone}
\begin{tabular}{@{}lccccccc@{}}
\toprule
Backbone & Avg. Image AUC & Avg. Pixel AUC & Wavelet Type & Level & Subbands & Sigma & Cov. Reg. \\
\midrule
efficientnet-b6 & 0.9884 & 0.8609 & haar & 1 & \texttt{LL} & 2.0 & 0.10 \\
efficientnet-b1 & 0.9851 & 0.7583 & sym4 & 1 & \texttt{LL} & 2.0 & 0.10 \\
efficientnet-b5 & 0.9851 & 0.8377 & haar & 2 & \texttt{LL} & 4.0 & 0.10 \\
efficientnet-b2 & 0.9848 & 0.8742 & haar & 1 & \texttt{LL} & 4.0 & 0.10 \\
efficientnet-b3 & 0.9847 & 0.8843 & haar & 1 & \texttt{LL} & 2.0 & 0.10 \\
efficientnet-b0 & 0.9844 & 0.7497 & sym4 & 1 & \texttt{LL} & 2.0 & 0.10 \\
efficientnet-b4 & 0.9818 & 0.7596 & sym4 & 1 & \texttt{LH\_LL} & 2.0 & 0.01 \\
resnet18 & 0.9681 & 0.7121 & sym4 & 2 & \texttt{LL} & 2.0 & 0.01 \\
\bottomrule
\end{tabular}
\end{table*}

\begin{table*}[h!]
\centering
\caption{Best average pixel AUC configuration by backbone Model.}
\label{tab:best_avg_pixel_auc_backbone}
\begin{tabular}{@{}lccccccc@{}}
\toprule
Backbone & Avg. Pixel AUC & Avg. Image AUC & Wavelet Type & Level & Subbands & Sigma & Cov. Reg. \\
\midrule
resnet18 & 0.9077 & 0.9430 & haar & 1 & \texttt{HH\_LH\_LL} & 2.0 & 0.001 \\
efficientnet-b3 & 0.8999 & 0.9777 & haar & 1 & \texttt{HH\_HL\_LH\_LL} & 2.0 & 0.100 \\
efficientnet-b1 & 0.8971 & 0.9741 & haar & 1 & \texttt{HH\_HL\_LH\_LL} & 2.0 & 0.100 \\
efficientnet-b2 & 0.8954 & 0.9805 & haar & 1 & \texttt{HH\_HL\_LH\_LL} & 2.0 & 0.100 \\
efficientnet-b4 & 0.8930 & 0.9774 & haar & 1 & \texttt{HH\_HL\_LH\_LL} & 2.0 & 0.100 \\
efficientnet-b0 & 0.8903 & 0.9702 & haar & 1 & \texttt{HH\_HL\_LH\_LL} & 2.0 & 0.100 \\
efficientnet-b6 & 0.8669 & 0.9859 & haar & 1 & \texttt{HL\_LH\_LL} & 2.0 & 0.100 \\
efficientnet-b5 & 0.8647 & 0.9794 & haar & 2 & \texttt{HH\_HL\_LH\_LL} & 4.0 & 0.100 \\
\bottomrule
\end{tabular}
\end{table*}

%====================
\subsection{Impact of Wavelet Configuration}
\label{subsec:results_wavelet_impact}

The choice of wavelet type, decomposition level, and selected subbands significantly influences the performance of WE-PaDiM.

\textbf{Impact of Wavelet Type:} Our experiments explored several wavelet families, primarily Haar, Daubechies (db2, db4), and Symlet (sym4). Analyzing the top-performing configurations across different backbones and optimization metrics (Tables \ref{tab:best_avg_image_auc_backbone} and \ref{tab:best_avg_pixel_auc_backbone}) reveals that both simpler (Haar) and more complex (sym4) wavelets can yield optimal results depending on the context. For instance, `efficientnet-b6' achieved its best Image AUC with the Haar wavelet, while `efficientnet-b1' used sym4 for its top Image AUC. For Pixel AUC, Haar was prevalent in many top configurations, particularly with ResNet-18 and several EfficientNet variants. This suggests that the simpler, more localized frequency separation of Haar might be beneficial for precise defect localization in some cases, whereas the smoother basis functions of Symlets might capture broader contextual features useful for overall image assessment. The optimal wavelet choice appears to be interconnected with the backbone architecture and the 
specific anomaly characteristics targeted.

\textbf{Impact of Subband Selection:} Table \ref{tab:subband_stats_overall} summarizes the overall performance statistics for different subband combinations when averaged across all tested backbones and classes. The results indicate that the \texttt{LL} (approximation) subband generally yields the highest average Image AUC (0.9821). For Pixel AUC, combinations incorporating more detail subbands, such as \texttt{HL\_LH\_LL} (0.8567) and \texttt{HH\_HL\_LL} (0.8529), tend to perform better on average. This suggests that low-frequency information is crucial for robust image-level detection, while high-frequency details contribute significantly to precise anomaly localization. The dimensionality $D_W$ is directly affected by the number of selected subbands; for instance, using only \texttt{LL} ($|S|=1$) offers the most significant reduction if all layers are considered. The per-class optimal subband configurations (Appendices \ref{app:per_class_image_auc}, \ref{app:per_class_pixel_auc}) further highlight that different MVTec AD categories benefit from different subband information, likely due to the varying nature of anomalies.

%====================
\subsection{Component Impact Analysis}

\begin{figure}[t]
    \centering
    \includegraphics[width=\linewidth]{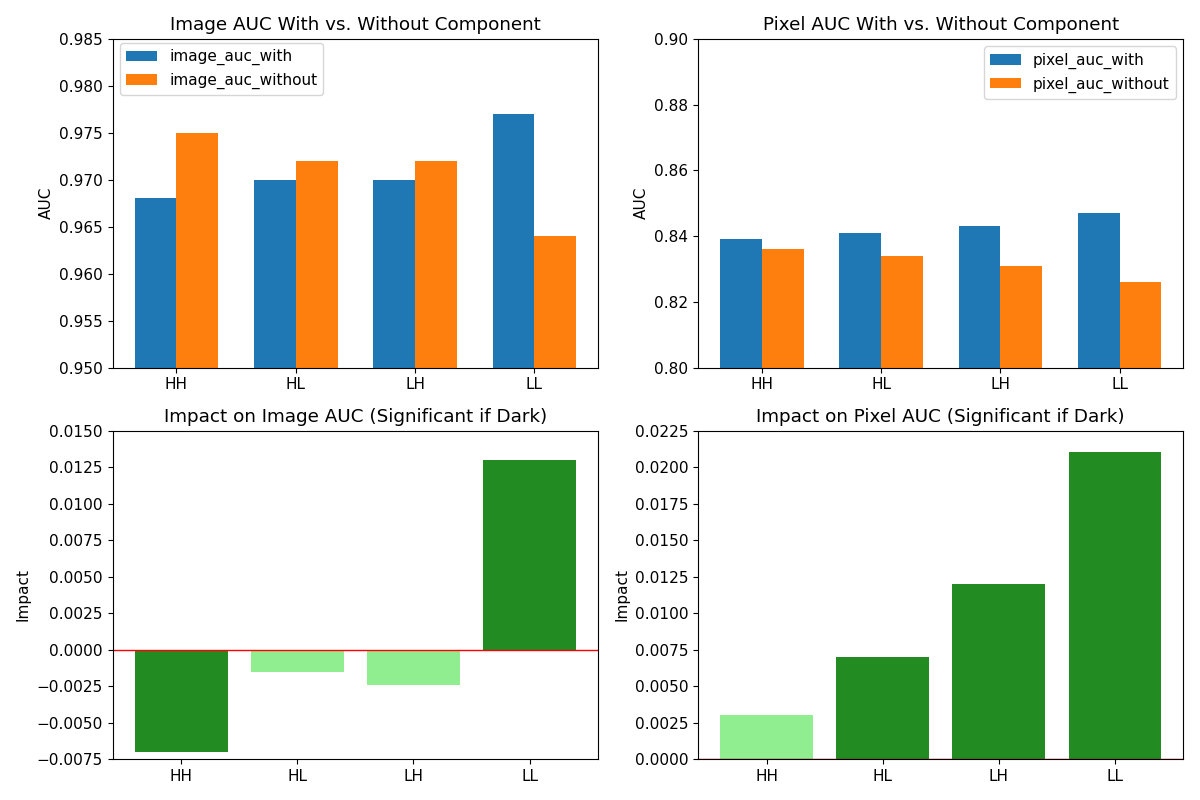}
    \caption{Impact of wavelet components on anomaly detection performance. Top row: Comparison of average Image AUC (left) and Pixel AUC (right) with and without each component. Bottom row: Differential impact of each component on Image AUC (left) and Pixel AUC (right), with dark green bars indicating statistically significant differences ($p<0.05$).}
    \label{fig:component_impact}
\end{figure}

To precisely quantify the contribution of each wavelet subband component, we conducted a statistical analysis comparing performance with and without specific components across all configurations. Fig.~\ref{fig:component_impact} illustrates the differential impact of each wavelet component (HH, HL, LH, and LL) on both image-level and pixel-level metrics.

The approximation component (LL) demonstrates the strongest positive impact on both Image AUC ($+0.0144$, $p<0.0001$) and Pixel AUC ($+0.0228$, $p<0.0001$). This substantial and statistically significant contribution explains why configurations using only the LL subband achieve strong image-level detection performance.

For detail components, we observe distinct patterns: the HH (diagonal detail) component shows a small but statistically significant negative impact on Image AUC ($-0.0050$, $p<0.0001$), suggesting its inclusion may introduce noise for overall classification. However, the HL (vertical detail) and LH (horizontal detail) components contribute positively to Pixel AUC ($+0.0092$ and $+0.0133$ respectively, both $p<0.0001$), confirming their importance for precise anomaly localization.

These quantitative findings support our configuration recommendations: (1) for maximizing image-level detection, prioritize the LL component; (2) for enhancing pixel-level localization, incorporate HL and LH detail components; and (3) when computational efficiency is a concern, the LL component offers the best individual performance-to-complexity ratio.

%====================
\subsection{Class-Specific Performance Analysis}

\begin{figure*}[t]
    \centering
    \includegraphics[width=\textwidth]{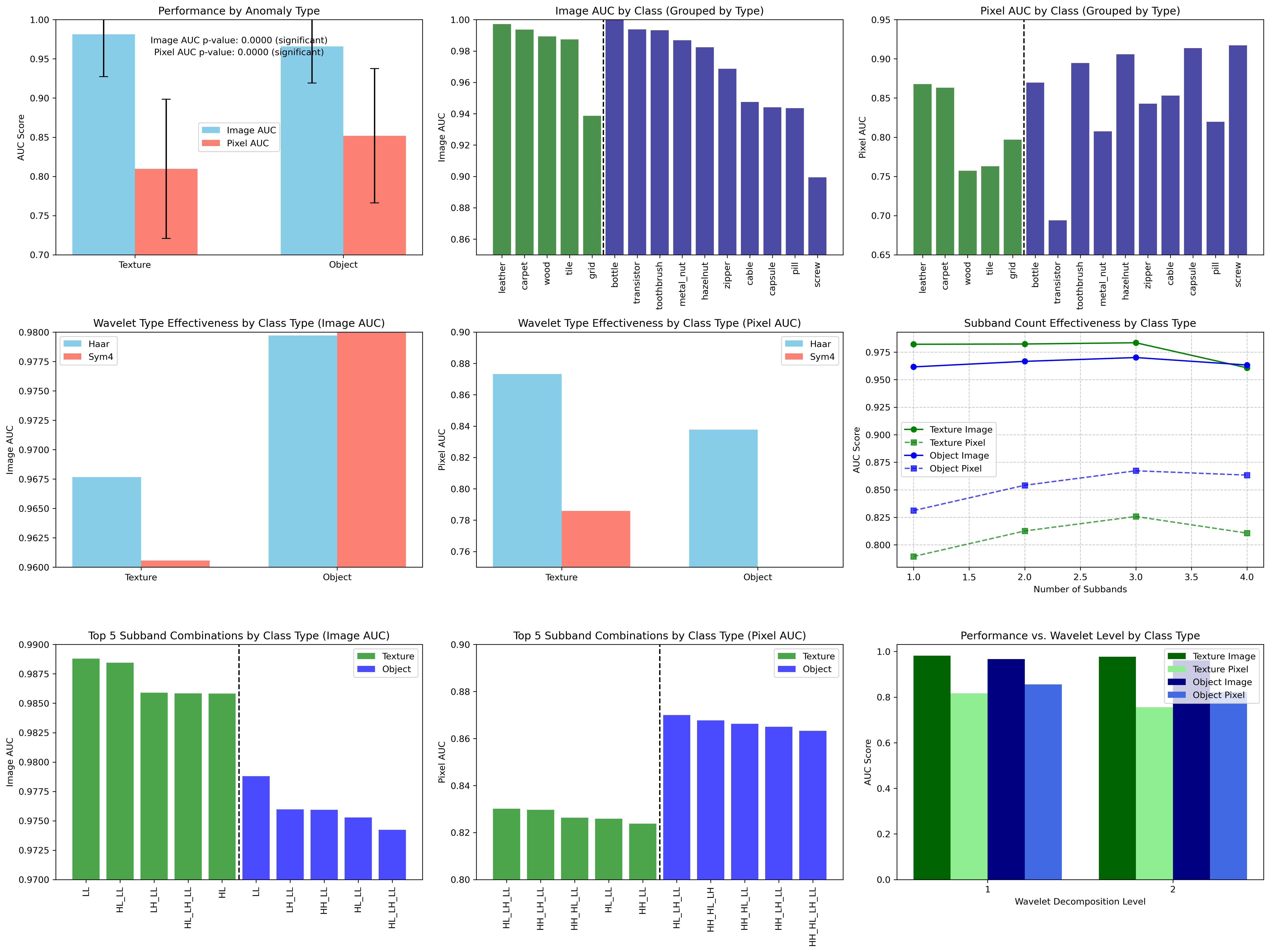}
    \caption{Class-specific performance analysis. Top row: Performance comparison between texture and object anomalies (left) and per-class performance for Image AUC (middle) and Pixel AUC (right). Middle row: Wavelet type effectiveness by class type for Image AUC (left) and Pixel AUC (middle), and subband count impact (right). Bottom row: Top 5 subband combinations for Image AUC (left) and Pixel AUC (middle), and wavelet level comparison (right).}
    \label{fig:class_analysis}
\end{figure*}

The MVTec AD dataset contains both texture categories (carpet, grid, leather, tile, wood) and object categories (bottle, cable, capsule, hazelnut, metal\_nut, pill, screw, toothbrush, transistor, zipper). Our analysis reveals statistically significant performance differences between these anomaly types (Fig.~\ref{fig:class_analysis}), with implications for configuration optimization.

Texture anomalies achieved higher average Image AUC ($0.980 \pm 0.019$) but significantly lower Pixel AUC ($0.812 \pm 0.079$) compared to object anomalies (Image AUC: $0.969 \pm 0.028$, Pixel AUC: $0.852 \pm 0.071$), with $p<0.0001$ for both metrics. This pattern suggests different optimal configurations for these anomaly types.

Wavelet type selection shows particularly strong class-specific patterns. For Image AUC on texture categories, Haar wavelets outperform Sym4, while for object categories, Sym4 wavelets yield better results. For Pixel AUC, Haar wavelets outperform Sym4 across both category types, but with a larger margin for textures. This suggests that simpler wavelets better capture the regular patterns in textures, while smoother wavelets may better represent object structures for classification.

The impact of subband count also varies by anomaly type. For texture categories, Image AUC remains consistently high regardless of subband count, while Pixel AUC shows modest improvement with additional subbands. In contrast, object categories show small improvements in Image AUC and substantial gains in Pixel AUC as more subbands are included, with the most pronounced improvements when moving from 1 to 2 subbands.

Notably, the optimal subband combinations differ between anomaly types. For texture categories, the LL subband alone achieves optimal Image AUC, while object categories benefit from various combinations including detail subbands (particularly HL\_LL). For Pixel AUC, texture categories perform best with HL\_LH\_LL combinations, while object categories show optimal performance with HH\_HL\_LH\_LL and similar combinations incorporating high-frequency components.

Wavelet decomposition level analysis reveals that Level 1 is generally optimal for Pixel AUC across both category types, while for Image AUC, texture categories perform equally well with both levels, and object categories show some preference for Level 1. This underscores that finer frequency decomposition is particularly important for localizing object anomalies.

These class-specific patterns suggest that practitioners should consider the nature of their anomaly detection task when configuring WE-PaDiM. For applications dominated by texture anomalies, simpler configurations with Haar wavelets and fewer subbands may suffice, while object anomaly detection benefits from more comprehensive configurations incorporating multiple detail subbands.

%====================
\subsection{Impact of Backbone Model and Other Parameters}
\label{subsec:results_backbone_params}

\textbf{Backbone Comparison:} The choice of CNN backbone significantly impacts WE-PaDiM's performance. Tables \ref{tab:best_avg_image_auc_backbone} and \ref{tab:best_avg_pixel_auc_backbone} show that various EfficientNet models (B0-B6) and ResNet-18 achieve competitive results, but their optimal wavelet configurations and peak performance differ. For Image AUC, EfficientNet variants generally dominate the top ranks, with `efficientnet-b6' (0.9884 Image AUC) leading. For Pixel AUC, `resnet18' (0.9077 Pixel AUC) demonstrated the strongest average performance. This indicates that deeper or wider EfficientNet architectures might be more adept at learning discriminative global features for image-level classification when enhanced with wavelet analysis focusing on approximation (`LL') bands. Conversely, the feature structure of ResNet-18, when combined with wavelet detail subbands, appears highly effective for localization tasks. The optimal WE-PaDiM configuration is thus not universal but rather dependent on the interplay between the backbone's feature representation and the specific anomaly characteristics being targeted by the wavelet decomposition.

\textbf{Impact of Sigma ($\sigma$) and Covariance Regularization ($\epsilon$):} The Gaussian smoothing sigma ($\sigma$) applied to the final anomaly map and the covariance regularization epsilon ($\epsilon$, or cov\_reg) are important hyperparameters. From the top configurations identified in Tables \ref{tab:best_avg_image_auc_backbone} and \ref{tab:best_avg_pixel_auc_backbone}, sigma values of 2.0 and 4.0 were frequently optimal. A sigma of 2.0 appeared more often in configurations leading to top Pixel AUC (e.g., for ResNet18, EfficientNet-B0/B1/B2/B3/B4/B6), suggesting that less aggressive smoothing preserves finer localization details. For Image AUC, both 2.0 and 4.0 were seen.
The covariance regularization $\epsilon$ commonly performed well at values of 0.10, 0.01, or 0.001. For instance, \texttt{efficientnet-b6} (top Image AUC) used $\epsilon=0.10$, while \texttt{resnet18} (top Pixel AUC) used $\epsilon=0.001$. The choice of $\epsilon$ influences the conditioning of the covariance matrices and thus the sensitivity of the Mahalanobis distance calculation.

\textbf{Impact of Wavelet Level:} Our experiments primarily focused on DWT level $J=1$. However, some configurations explored $J=2$. As seen in Table \ref{tab:best_avg_image_auc_backbone}, `efficientnet-b5' and `resnet18' achieved their best average Image AUC using level $J=2$ with the \texttt{LL} subband. This implies that for certain backbones, a coarser approximation subband from a higher decomposition level can be more effective for image-level classification. For instance, with `efficientnet-b5`, using $J=2$ (`LL' subband) resulted in an Image AUC of 0.9851, while its best $J=1$ configuration (e.g., with \texttt{HH\_HL\_LH\_LL} for Pixel AUC) yielded a slightly lower Image AUC of 0.9794. The choice of level $J$ interacts with the spatial resolution of the feature maps and the scale of anomalies, offering another dimension for tuning. Generally, $J=1$ was prevalent in most top configurations, particularly for Pixel AUC optimization (Table \ref{tab:best_avg_pixel_auc_backbone}), suggesting that finer details from the first level of decomposition are often crucial for localization.

\textbf{Impact of Subband Selection:} Table \ref{tab:subband_stats_overall} summarizes the overall performance statistics for different subband combinations when averaged across all tested backbones and classes. The results indicate that the \texttt{LL} (approximation) subband generally yields the highest average Image AUC (0.9821). For Pixel AUC, combinations incorporating more detail subbands, such as \texttt{HL\_LH\_LL} (0.8567) and \texttt{HH\_HL\_LL} (0.8529), tend to perform better on average. This suggests that low-frequency information is crucial for robust image-level detection, while high-frequency details contribute significantly to precise anomaly localization. The dimensionality $D_W$ is directly affected by the number of selected subbands; for instance, using only \texttt{LL} ($|S|=1$) offers the most significant reduction if all layers are considered.

\begin{table}[h!]
\centering
\caption{Overall performance statistics by subband combination (averaged across all backbones and classes).}
\label{tab:subband_stats_overall}
\begin{tabular}{@{}lcccccc@{}}
\toprule
& \multicolumn{2}{c}{Image AUC} & \multicolumn{2}{c}{Pixel AUC} \\
\midrule
Subbands & Mean & Std  & Mean & Std  \\
\midrule
\texttt{LL} & 0.9821 & 0.0236 & 0.8352 & 0.0889 \\
\texttt{HL\_LL} & 0.9797 & 0.0274  & 0.8471 & 0.0846 \\
\texttt{LH\_LL} & 0.9793 & 0.0272  & 0.8466 & 0.0834 \\
\texttt{HH\_LL} & 0.9791 & 0.0275  & 0.8443 & 0.0876 \\
\texttt{HL\_LH\_LL} & 0.9781 & 0.0318  & 0.8567 & 0.0795 \\
\texttt{HH\_HL\_LL} & 0.9772 & 0.0301  & 0.8529 & 0.0853 \\
\texttt{HH\_LH\_LL} & 0.9762 & 0.0355 & 0.8532 & 0.0813 \\
\texttt{HL} & 0.9684 & 0.0425  & 0.8081 & 0.0922 \\
\texttt{HL\_LH} & 0.9681 & 0.0462  & 0.8445 & 0.0774  \\
\texttt{LH} & 0.9670 & 0.0388  & 0.8179 & 0.0807  \\
\texttt{HH\_HL\_LH} & 0.9665 & 0.0459  & 0.8506 & 0.0764  \\
\texttt{HH\_HL} & 0.9627 & 0.0490  & 0.8262 & 0.0860  \\
\texttt{HH\_HL\_LH\_LL} & 0.9624 & 0.1266  & 0.8457 & 0.1330  \\
\texttt{HH\_LH} & 0.9620 & 0.0476  & 0.8322 & 0.0791 \\
\texttt{HH} & 0.9562 & 0.0532  & 0.8075 & 0.0879  \\
\bottomrule
\end{tabular}
\end{table}

%====================
\subsection{Per-Class Results and Top Configurations}
\label{subsec:results_per_class}

Detailed per-class results, showing the consolidated best performing configurations for each MVTec AD category ranked primarily by Image AUC (with Pixel AUC as a tie-breaker), are provided in Appendix \ref{app:per_class_image_auc} (Table \ref{tab:best_per_class_image_auc_consolidated}). Similarly, Appendix \ref{app:per_class_pixel_auc} (Table \ref{tab:best_per_class_pixel_auc_consolidated}) presents per-class results optimized primarily for Pixel AUC. These tables highlight the variability in optimal parameters across different object and texture categories. For example, several classes such as 'bottle', 'carpet', and 'leather' achieve a perfect Image AUC of 1.0000 (see Table \ref{tab:best_per_class_image_auc_consolidated}). For Pixel AUC, the 'screw' class reached a high of 0.9853 with a ResNet-18 backbone (see Table \ref{tab:best_per_class_pixel_auc_consolidated}).

%====================
\section{Discussion}
\label{sec:discussion}

The experimental results demonstrate that Wavelet-Enhanced PaDiM (WE-PaDiM) is a promising approach for industrial anomaly detection and localization, achieving competitive performance on the MVTec AD dataset. The core "DWT-before-concatenation" strategy, where wavelet transforms are applied individually to multi-layer CNN features before subband selection and channel-wise concatenation, offers a principled alternative to PaDiM's original random feature selection.

\textbf{Significance of Wavelet-Based Feature Structuring:}
Our findings support the hypothesis that anomalies often possess distinct frequency characteristics that can be effectively captured using DWT. By decomposing feature maps into different frequency subbands, WE-PaDiM allows for a more targeted feature representation. For instance, the consistent high performance of the \texttt{LL} (approximation) subband for Image AUC across various backbones (Table \ref{tab:subband_stats_overall}) suggests that coarse, low-frequency information from CNN features is highly discriminative for overall image classification. This aligns with the intuition that many global anomalies or significant structural deviations might be well-represented in these approximation coefficients.

Conversely, for Pixel AUC, which measures localization accuracy, combinations including detail subbands (\texttt{LH}, \texttt{HL}, \texttt{HH}) often yielded better results (Table \ref{tab:subband_stats_overall}). For example, \texttt{HL\_LH\_LL} and \texttt{HH\_HL\_LH\_LL} showed strong average Pixel AUCs. This is logical, as subtle anomalies like scratches, texture defects, or edge imperfections are typically characterized by high-frequency components captured in the detail subbands. The per-class results in Appendices \ref{app:per_class_image_auc} and \ref{app:per_class_pixel_auc} further illustrate this: categories with fine-grained defects may benefit more from configurations emphasizing detail subbands. This structured selection based on frequency content provides a degree of interpretability not present in random selection, allowing for a more informed choice of features based on expected anomaly types.

\textbf{Performance Trade-offs and Configuration Choices:}
A key observation is the trade-off between Image AUC and Pixel AUC based on wavelet configuration and backbone. As seen in Tables \ref{tab:best_avg_image_auc_backbone} and \ref{tab:best_avg_pixel_auc_backbone}, configurations excelling at image-level detection (often using primarily \texttt{LL} bands from deeper EfficientNet models) might not be the absolute best for pixel-level localization, and vice-versa (where ResNet-18 with combined subbands performed well for Pixel AUC). This suggests that the optimal WE-PaDiM setup can be tailored to the specific requirements of an industrial inspection task – prioritizing high detection rates or precise localization.

The choice of wavelet type (e.g., Haar vs. Symlet) and level also plays a role. While Haar wavelets, being computationally simpler and more localized, frequently appeared in top pixel-level configurations, smoother wavelets like Symlet4 were present in some top image-level configurations, particularly with certain EfficientNet backbones. Using $J=2$ decomposition was beneficial for Image AUC with `efficientnet-b5` and `resnet18` when using only the \texttt{LL} band, indicating that an even coarser approximation can sometimes be optimal for global assessment.

\textbf{Comparison with Original PaDiM and Dimensionality:}
While a direct, controlled comparison with the original PaDiM's random selection (tuned for optimal $D_R$) was not the primary focus of the presented results, the high AUC values achieved by WE-PaDiM (e.g., average 99.32\% Image AUC) are competitive with state-of-the-art methods, including PaDiM. The dimension $D_W$ in WE-PaDiM is determined by $|S| \sum C_l$. Using only \texttt{LL} ($|S|=1$) leads to a dimensionality equal to the sum of channels of selected layers, which is already a reduction compared to PaDiM's initial concatenation before random selection if the same layers are used. Using more subbands (e.g., $|S|=3$ for \texttt{LL,LH,HL}) increases $D_W$, but this is a structured increase based on frequency information rather than random sampling. Future work could involve a detailed comparison of $D_W$ vs. PaDiM's $D_R$ for similar performance levels.

\textbf{Interpretability and Practical Implications:}
The ability to select specific subbands offers a step towards more interpretable feature selection. For example, if a particular industrial process is prone to high-frequency scratch-like defects, a WE-PaDiM configuration emphasizing \texttt{LH}, \texttt{HL}, and/or \texttt{HH} subbands might be chosen. This is a more guided approach than random selection. The comparable computational efficiency (as suggested by the abstract, though detailed timings are pending) makes WE-PaDiM a practical option.

\textbf{Limitations and Future Work:}
The current study primarily explored $J=1$ and $J=2$ DWT levels and a subset of wavelet families. A more exhaustive search, including higher decomposition levels or other wavelet types (e.g., Coiflets, Biorthogonal), could yield further insights. The interaction between wavelet parameters and specific CNN layer choices also warrants deeper investigation; it's possible that different layers benefit from different wavelet decompositions. While the "DWT-before-concatenation" approach proved effective, alternative strategies for integrating DWT (e.g., DWT after concatenation, or adaptive wavelet transforms) could be explored. Finally, formally quantifying the interpretability gains and conducting statistical significance tests for performance differences between configurations would further strengthen the findings. The role of adaptive fusion, for which infrastructure exists in the codebase, could also be more deeply explored as a potential enhancement.

%====================
\section{Conclusion}
\label{sec:conclusion}

In this paper, we introduced Wavelet-Enhanced PaDiM (WE-PaDiM), a novel approach for unsupervised anomaly detection and localization in industrial images. By integrating Discrete Wavelet Transform analysis directly into the feature processing pipeline of PaDiM—applying DWT individually to multi-layer CNN feature maps and then selecting and concatenating specific subband coefficients—WE-PaDiM offers a structured and interpretable alternative to random feature selection.

Our comprehensive evaluations on the MVTec AD dataset, using a range of CNN backbones including ResNet-18 and EfficientNet variants (B0-B6), demonstrate the efficacy of this "DWT-before-concatenation" strategy. WE-PaDiM achieves high performance, with average Image AUC reaching 99.32\% and average Pixel AUC reaching 92.10\% when configurations are optimized per class. We found that the choice of wavelet type, decomposition level, and particularly the selected subbands significantly impacts performance, revealing trade-offs between image-level detection (often favoring \texttt{LL} subbands) and pixel-level localization (often benefiting from combinations of \texttt{LL}, \texttt{LH}, and \texttt{HL} subbands). This allows for tailoring the method to specific industrial inspection needs.

WE-PaDiM provides a competitive method that not only matches the performance of strong baselines but also enhances the feature selection process with frequency-domain insights, potentially leading to better understanding and control over which image characteristics are prioritized for anomaly detection. Future work may involve exploring a wider range of wavelet configurations, deeper analysis of layer-wavelet interactions, and alternative DWT integration strategies. Overall, WE-PaDiM presents a valuable contribution to the field of deep learning-based industrial anomaly detection.

% ==============================
\appendix
\section{Per-Class Results Optimized for Image AUC}
\label{app:per_class_image_auc}
Table \ref{tab:best_per_class_image_auc_consolidated} shows the best performing configurations for each MVTec AD class, where the primary sorting criterion is Image AUC. If multiple configurations achieve the same top Image AUC for a class, the one with the highest Pixel AUC among them is prioritized for listing the specific AUC values shown, and parameters reflect all unique values across tied top configurations.

\begin{table*}[ht!]
\centering
\caption{Best Performing Configurations for Each MVTec AD Class (Optimized for Image AUC, then Pixel AUC for ties). Parameters listed are unique values from configurations achieving the top Image AUC. Eff. = EfficientNet.}
\label{tab:best_per_class_image_auc_consolidated}
\begin{adjustbox}{width=\textwidth,center}
\begin{tabular}{@{}lcccccccccr@{}}
\toprule
Class & Image AUC & Pixel AUC & Backbone(s) & Wavelet Type(s) & Level(s) & Subband(s) & Sigma(s) & Cov.Reg.(s) & Opt. & \# Cfgs \\
\midrule
bottle & 1.0000 & 0.9393 & Eff-b5, resnet18, Eff-b4, ... & haar, sym4 & 1, 2 & \texttt{HH\_LL}, \texttt{LL}, ... & 4.0, 2.0 & 0.001, 0.01, 0.1 & image, pixel & 430 \\
carpet & 1.0000 & 0.9393 & Eff-b0, Eff-b1, Eff-b3, ... & sym4, haar & 1, 2 & \texttt{HL}, \texttt{HH\_HL}, ... & 2.0, 4.0 & 0.1, 0.01, 0.001 & image, pixel & 236 \\
hazelnut & 1.0000 & 0.9271 & Eff-b4, Eff-b1, Eff-b6, ... & sym4, haar & 1 & \texttt{LH\_LL}, \texttt{LL}, ... & 2.0, 4.0 & 0.01, 0.1 & image & 22 \\
leather & 1.0000 & 0.9628 & Eff-b0, Eff-b1, Eff-b3, ... & sym4, haar & 1, 2 & \texttt{LH}, \texttt{LL}, ... & 2.0, 4.0 & 0.1, 0.01, 0.001 & image, pixel & 386 \\
toothbrush & 1.0000 & 0.9362 & Eff-b5, resnet18, Eff-b4, ... & haar, sym4 & 1, 2 & \texttt{HL\_LH}, \texttt{LL}, ... & 4.0, 2.0 & 0.001, 0.01, 0.1 & image, pixel & 252 \\
transistor & 1.0000 & 0.7407 & Eff-b4, Eff-b1, Eff-b6, ... & sym4, haar & 1 & \texttt{LH\_LL}, \texttt{HH\_HL\_LL}, ... & 2.0, 4.0 & 0.01, 0.1, 0.001 & image, pixel & 48 \\
tile & 0.9993 & 0.7030 & Eff-b1 & sym4 & 1 & \texttt{HH} & 2.0 & 0.1 & image & 2 \\
metal\_nut & 0.9990 & 0.8778 & Eff-b4, Eff-b5 & sym4, haar & 1 & \texttt{HH\_LL}, \texttt{HH\_HL\_LH\_LL}, ... & 2.0, 4.0 & 0.01, 0.001 & image & 10 \\
wood & 0.9982 & 0.8538 & Eff-b1, resnet18 & sym4, haar & 1 & \texttt{HH\_HL\_LH\_LL}, ... & 2.0 & 0.1, 0.001 & image, pixel & 5 \\
zipper & 0.9955 & 0.7047 & Eff-b1 & sym4 & 1 & \texttt{LL} & 2.0 & 0.1 & image & 2 \\
cable & 0.9948 & 0.8046 & Eff-b5 & haar & 2 & \texttt{LL} & 4.0 & 0.1 & pixel & 2 \\
grid & 0.9900 & 0.8013 & Eff-b0, Eff-b5 & sym4, haar & 1, 2 & \texttt{HL}, \texttt{HL\_LH\_LL} & 2.0, 4.0 & 0.1 & image, pixel & 4 \\
capsule & 0.9852 & 0.9315 & Eff-b6 & haar & 1 & \texttt{HL} & 4.0 & 0.01 & image & 2 \\
pill & 0.9804 & 0.8273 & Eff-b0 & haar & 1 & \texttt{LL} & 2.0 & 0.1 & pixel & 2 \\
screw & 0.9553 & 0.9217 & Eff-b5 & haar & 2 & \texttt{HL\_LL} & 4.0 & 0.1 & pixel & 2 \\
\midrule
\textbf{Average} & \textbf{0.9932} & \textbf{0.8383} & \multicolumn{7}{l}{\textit{(Average of best Image AUCs; corresponding Pixel AUCs then averaged)}} &  \\
\bottomrule
\end{tabular}
\end{adjustbox}
\end{table*}

\section{Per-Class Results Optimized for Pixel AUC}
\label{app:per_class_pixel_auc}
Table \ref{tab:best_per_class_pixel_auc_consolidated} shows the best performing configurations for each MVTec AD class, where the primary sorting criterion is Pixel AUC. If multiple configurations achieve the same top Pixel AUC for a class, the one with the highest Image AUC among them is prioritized for listing the specific AUC values shown, and parameters reflect all unique values across tied top configurations.

\begin{table*}[ht!]
\centering
\caption{Best Performing Configurations for Each MVTec AD Class (Optimized for Pixel AUC, then Image AUC for ties). Parameters listed are unique values from configurations achieving the top Pixel AUC. Eff. = EfficientNet.}
\label{tab:best_per_class_pixel_auc_consolidated}
\begin{adjustbox}{width=\textwidth,center}
\begin{tabular}{@{}lcccccccccr@{}}
\toprule
Class & Pixel AUC & Image AUC & Backbone(s) & Wavelet Type(s) & Level(s) & Subband(s) & Sigma(s) & Cov.Reg.(s) & Opt. & \# Cfgs \\
\midrule
screw & 0.9853 & 0.8469 & resnet18 & haar & 1 & \texttt{LH\_LL} & 2.0 & 0.001 & pixel & 1 \\
leather & 0.9628 & 1.0000 & Eff-b3 & haar & 1 & \texttt{HH\_LL} & 2.0 & 0.1 & pixel & 2 \\
capsule & 0.9574 & 0.8732 & resnet18 & haar & 1 & \texttt{HH\_HL\_LH\_LL} & 2.0 & 0.001 & pixel & 1 \\
hazelnut & 0.9541 & 0.9768 & resnet18 & haar & 1 & \texttt{HH\_HL\_LL} & 2.0 & 0.001 & pixel & 1 \\
cable & 0.9469 & 0.9076 & resnet18 & haar & 1 & \texttt{LH\_LL} & 2.0 & 0.001 & pixel & 1 \\
toothbrush & 0.9405 & 0.9611 & Eff-b4 & haar & 1 & \texttt{HL\_LH} & 2.0 & 0.1 & pixel & 2 \\
bottle & 0.9393 & 1.0000 & resnet18 & haar & 1 & \texttt{HH\_HL\_LH\_LL} & 2.0 & 0.001 & pixel & 1 \\
carpet & 0.9393 & 1.0000 & Eff-b3 & haar & 1 & \texttt{HH\_LL} & 2.0 & 0.1 & pixel & 2 \\
grid & 0.9347 & 0.9006 & resnet18 & haar & 1 & \texttt{LH\_LL} & 2.0 & 0.001 & pixel & 1 \\
pill & 0.9239 & 0.9714 & resnet18 & haar & 1 & \texttt{LL} & 2.0 & 0.001 & pixel & 1 \\
zipper & 0.9096 & 0.9034 & resnet18 & haar & 1 & \texttt{HH\_LH\_LL} & 2.0 & 0.001 & pixel & 1 \\
metal\_nut & 0.9095 & 0.9956 & Eff-b2 & haar & 1 & \texttt{HH\_LH\_LL} & 4.0 & 0.1 & image & 2 \\
wood & 0.8619 & 0.9965 & resnet18 & haar & 1 & \texttt{HH\_LH\_LL} & 2.0 & 0.001 & pixel & 1 \\
tile & 0.8544 & 0.9899 & resnet18 & haar & 1 & \texttt{LL} & 2.0 & 0.001 & pixel & 1 \\
transistor & 0.7961 & 0.9950 & Eff-b4 & sym4 & 1 & \texttt{HL\_LH\_LL} & 2.0 & 0.01 & image & 2 \\
\midrule
\textbf{Average} & \textbf{0.9210} & \textbf{0.9555} & \multicolumn{7}{l}{\textit{(Average of best Pixel AUCs; corresponding Image AUCs then averaged)}} &  \\
\bottomrule
\end{tabular}
\end{adjustbox}
\end{table*}

% ==============================
\section*{Acknowledgments}
This research was supported by the Technology Innovation Program (ATC+ Program, Project No. 20014131, ``25nm X-ray Inspection System for Semiconductor Backend Process'') funded by the Ministry of Trade, Industry and Energy (MOTIE, Korea). Additional support for C.G. and T.A. was provided by the National Science Foundation (NSF) under Grant No. 2430236, and B.M. was supported by the Faculty Research Fund of Sejong University.

% ==============================
\section*{Code Availability}
The implementation of WE-PaDiM is available at \url{https://github.com/BioHPC/WE-PaDiM}.

% ==============================
\bibliographystyle{unsrtnat} % plain numeric, unbranded
\bibliography{references}

\begin{thebibliography}{22}
\providecommand{\natexlab}[1]{#1}
\providecommand{\url}[1]{\texttt{#1}}
\expandafter\ifx\csname urlstyle\endcsname\relax
  \providecommand{\doi}[1]{doi: #1}\else
  \providecommand{\doi}{doi: \begingroup \urlstyle{rm}\Url}\fi

\bibitem[Bergmann et~al.(2019)Bergmann, Fauser, Sattlegger, and
  Steger]{bergmann2019mvtec}
Paul Bergmann, Michael Fauser, David Sattlegger, and Carsten Steger.
\newblock Mvtec ad--a comprehensive real-world dataset for unsupervised anomaly
  detection.
\newblock \emph{Proceedings of the IEEE/CVF Conference on Computer Vision and
  Pattern Recognition}, pages 9592--9600, 2019.
\newblock \doi{10.1109/CVPR.2019.00982}.

\bibitem[Defard et~al.(2020)Defard, Setkov, Loesch, and Audigier]{padim2020}
Thomas Defard, Aleksandr Setkov, Angelique Loesch, and Romaric Audigier.
\newblock Padim: A patch distribution modeling framework for anomaly detection
  and localization.
\newblock In \emph{Proceedings of the International Conference on Pattern
  Recognition Workshops (ICPR Workshops)}, pages 475--489, 2020.
\newblock \doi{10.48550/arXiv.2011.08785}.

\bibitem[Mallat(1989)]{mallat1989wavelet}
St{'e}phane Mallat.
\newblock A theory for multiresolution signal decomposition: The wavelet
  representation.
\newblock \emph{IEEE Transactions on Pattern Analysis and Machine
  Intelligence}, 11\penalty0 (7):\penalty0 674--693, 1989.

\bibitem[He et~al.(2016)He, Zhang, Ren, and Sun]{he2016resnet}
Kaiming He, Xiangyu Zhang, Shaoqing Ren, and Jian Sun.
\newblock Deep residual learning for image recognition.
\newblock In \emph{Proceedings of the IEEE Conference on Computer Vision and
  Pattern Recognition}, pages 770--778, 2016.

\bibitem[Tan and Le(2019)]{tan2019efficientnet}
Mingxing Tan and Quoc~V. Le.
\newblock Efficientnet: Rethinking model scaling for convolutional neural
  networks.
\newblock In \emph{Proceedings of the 36th International Conference on Machine
  Learning (ICML)}, volume~97 of \emph{Proceedings of Machine Learning
  Research}, pages 6105--6114, 2019.

\bibitem[Bergmann et~al.(2018)Bergmann, L{\"o}we, Fauser, Sattlegger, and
  Steger]{bergmann2018ssimAE}
Paul Bergmann, Sindy L{\"o}we, Michael Fauser, David Sattlegger, and Carsten
  Steger.
\newblock Improving unsupervised defect segmentation by applying structural
  similarity to autoencoders.
\newblock In \emph{arXiv preprint arXiv:1807.02011}, 2018.
\newblock URL \url{https://arxiv.org/abs/1807.02011}.

\bibitem[Gong et~al.(2019)Gong, Liu, Le, Saha, Mansour, Venkatesh, and van~den
  Hengel]{gong2019memae}
Dong Gong, Lingqiao Liu, Victoria Le, Budhaditya Saha, Moussa~Reda Mansour,
  Svetha Venkatesh, and Anton van~den Hengel.
\newblock Memorizing normality to detect anomaly: Memory‐augmented deep
  autoencoder for unsupervised anomaly detection.
\newblock In \emph{Proceedings of the IEEE/CVF International Conference on
  Computer Vision (ICCV)}, pages 1705--1714, 2019.

\bibitem[Schlegl et~al.(2017)Schlegl, Seeb{\"o}ck, Waldstein,
  Schmidt‐Erfurth, and Langs]{schlegl2017anogan}
Thomas Schlegl, Philipp Seeb{\"o}ck, Sebastian~M. Waldstein, Ursula
  Schmidt‐Erfurth, and Georg Langs.
\newblock Unsupervised anomaly detection with generative adversarial networks
  to guide marker discovery.
\newblock In \emph{Information Processing in Medical Imaging (IPMI)}, pages
  146--157, 2017.

\bibitem[Ak{\c{c}}ay et~al.(2018)Ak{\c{c}}ay, Atapour{-}Abarghouei, and
  Breckon]{akcay2018ganomaly}
Samet Ak{\c{c}}ay, Amir Atapour{-}Abarghouei, and Toby~P. Breckon.
\newblock {GANomaly}: Semi{\textendash}supervised anomaly detection via
  adversarial training.
\newblock In \emph{Asian Conference on Computer Vision (ACCV) Workshops}, pages
  622--637, 2018.

\bibitem[Cohen and Hoshen(2021)]{cohen2020spade}
Niv Cohen and Yedid Hoshen.
\newblock Sub-image anomaly detection with deep pyramid correspondences.
\newblock In \emph{International Conference on Learning Representations}, 2021.

\bibitem[Roth et~al.(2022)Roth, Pemula, Zepeda, Sch{\"o}lkopf, Brox, and
  Gehler]{roth2022patchcore}
Karsten Roth, Latha Pemula, Joaquin Zepeda, Bernhard Sch{\"o}lkopf, Thomas
  Brox, and Peter Gehler.
\newblock Towards total recall in industrial anomaly detection.
\newblock In \emph{Proceedings of the IEEE/CVF Conference on Computer Vision
  and Pattern Recognition}, pages 14318--14328, 2022.
\newblock \doi{10.48550/arXiv.2106.08265}.

\bibitem[Bergmann et~al.(2020)Bergmann, L{\"o}we, Fauser, Sattlegger, and
  Steger]{bergmann2020studentteacher}
Paul Bergmann, Sindy L{\"o}we, Michael Fauser, David Sattlegger, and Carsten
  Steger.
\newblock Uninformed students: Student–teacher anomaly detection with
  discriminative latent embeddings.
\newblock In \emph{Proceedings of the IEEE/CVF Conference on Computer Vision
  and Pattern Recognition Workshops (CVPRW)}, pages 190--191, 2020.

\bibitem[Li et~al.(2021)Li, Salzmann, and Fua]{li2021cutpaste}
Yang Li, Mathieu Salzmann, and Pascal Fua.
\newblock Cutpaste: Self-supervised learning for anomaly detection and
  localization.
\newblock In \emph{Proceedings of the IEEE/CVF Conference on Computer Vision
  and Pattern Recognition Workshops (CVPRW)}, pages 461--470, 2021.

\bibitem[Yu et~al.(2023)Yu, Su, Gong, Hong, Fu, and Li]{yu2023fastflow}
Tian Yu, Yongzhi Su, Boqing Gong, Lanqing Hong, Jianlong Fu, and Hongsheng Li.
\newblock Fastflow: Unsupervised anomaly detection and localization via 2d
  normalizing flows.
\newblock In \emph{Proceedings of the 31st ACM International Conference on
  Multimedia (ACM MM)}, pages 1229--1238, 2023.
\newblock \doi{10.1145/3581783.3612047}.

\bibitem[Zavrtanik et~al.(2021)Zavrtanik, Kristan, and
  Skocaj]{zavrtanik2021draem}
Vitja Zavrtanik, Marko Kristan, and Danijel Skocaj.
\newblock Draem: A discriminative reconstruction autoencoder for
  weakly-supervised anomaly detection.
\newblock In \emph{Proceedings of the IEEE/CVF International Conference on
  Computer Vision (ICCV)}, pages 833--842, 2021.

\bibitem[Ruff et~al.(2021)Ruff, Meadowcroft, Vandermeulen, and
  et~al.]{ruff2021differnet}
Lukas Ruff, Jacob Meadowcroft, Robert~A. Vandermeulen, and et~al.
\newblock Efficient one-class density estimation with support vector data
  description and normalizing flows.
\newblock In \emph{International Conference on Learning Representations
  (ICLR)}, 2021.
\newblock arXiv:2106.09694.

\bibitem[Ren and Liu(2022)]{ren2022lightpadim}
Chaoqiang Ren and Dengfeng Liu.
\newblock Light padim for unsupervised defect detection and location.
\newblock In \emph{Proceedings of the 8th International Conference on Computing
  and Artificial Intelligence (ICCAI)}, pages 739--744, 2022.
\newblock \doi{10.1145/3532213.3532326}.

\bibitem[Ibarra and Peeples(2025)]{ibarra2025padimace}
Angelina Ibarra and Joshua Peeples.
\newblock Padim-ace: Patch distribution modeling framework with adaptive cosine
  estimator for anomaly detection and localization in synthetic aperture radar
  imagery.
\newblock \emph{arXiv preprint arXiv:2504.08049}, 2025.

\bibitem[Mallat(2012)]{mallat2012wavelet}
St{\'e}phane Mallat.
\newblock Group invariant scattering.
\newblock \emph{Communications on Pure and Applied Mathematics}, 65\penalty0
  (10):\penalty0 1331--1398, 2012.

\bibitem[Kanarachos et~al.(2015)Kanarachos, Mathew, Chroneos, and
  Fitzpatrick]{kanarachos2015anomaly}
Stratis Kanarachos, Jino Mathew, Alexander Chroneos, and M~Fitzpatrick.
\newblock Anomaly detection in time series data using a combination of
  wavelets, neural networks and hilbert transform.
\newblock In \emph{2015 6th International Conference on Information,
  Intelligence, Systems and Applications (IISA)}, pages 1--6. IEEE, 2015.

\bibitem[Donoho(1995)]{donoho1995adapting}
David~L Donoho.
\newblock De-noising by soft-thresholding.
\newblock \emph{IEEE Transactions on Information Theory}, 41\penalty0
  (3):\penalty0 613--627, 1995.
\newblock \doi{10.1109/18.382009}.

\bibitem[Paszke et~al.(2019)Paszke, Gross, Massa, Lerer, Bradbury, Chanan,
  Killeen, Lin, Gimelshein, Antiga, Desmaison, Kopf, Yang, DeVito, Raison,
  Tejani, Chilamkurthy, Steiner, Fang, Bai, and Chaudhuri]{paszke2019pytorch}
Adam Paszke, Sam Gross, Francisco Massa, Adam Lerer, James Bradbury, Gregory
  Chanan, Trevor Killeen, Zeming Lin, Natalia Gimelshein, Luca Antiga, Alban
  Desmaison, Andreas Kopf, Edward Yang, Zachary DeVito, Martin Raison, Alykhan
  Tejani, Sasank Chilamkurthy, Benoit Steiner, Lu~Fang, Junjie Bai, and Soumith
  Chaudhuri.
\newblock Pytorch: An imperative style, high-performance deep learning library.
\newblock In \emph{Advances in Neural Information Processing Systems}, pages
  8024--8035, 2019.

\end{thebibliography}

\end{document}